\documentclass[runningheads]{llncs}
\usepackage{graphicx}
\usepackage{comment}
\usepackage{amsmath,amssymb} 
\usepackage{color}
\usepackage{multirow}
\usepackage{booktabs}
\usepackage{bbding}

\usepackage{array}

\newcolumntype{Y}{>{\centering\arraybackslash}m{1cm}}
\usepackage{footmisc}
\usepackage{threeparttable}
\usepackage{siunitx,booktabs,etoolbox}
\usepackage[colorlinks=true,citecolor=blue,urlcolor=black]{hyperref}
\usepackage[misc]{ifsym}

\providecommand{\mb}[1]{\mathbf{#1}}

\providecommand{\mbx}{\mb{x}}
\providecommand{\mby}{\mb{y}}
\providecommand{\mbz}{\mb{z}}

\providecommand{\mbT}{\mb{T}}

\makeatletter
\newcommand{\printfnsymbol}[1]{%
  \textsuperscript{\@fnsymbol{#1}}%
}
\makeatother

\newcommand\blfootnote[1]{%
  \begingroup
  \renewcommand\thefootnote{}\footnote{#1}%
  \addtocounter{footnote}{-1}%
  \endgroup
}

\usepackage{xcolor}

\begin{document}

\pagestyle{headings}
\mainmatter

\title{Improved post-hoc probability calibration for out-of-domain MRI segmentation} %

\titlerunning{Probability calibration for segmenting out-of-domain MRI}
%
\author{Cheng Ouyang\inst{1}$^($\Envelope$^)$, Shuo Wang\inst{2}, Chen Chen\inst{1}, Zeju Li\inst{1}, Wenjia Bai\inst{1,3,4}, Bernhard Kainz\inst{1,5} Daniel Rueckert\inst{1,6}}

\authorrunning{Ouyang et al.}
\institute{BioMedIA Group, Department of Computing, Imperial College London, UK
\and School of Basic Medical Sciences, Fudan University, China
\and Department of Brain Sciences, Imperial College London, UK
\and Data Science Institute, Imperial College London, UK
\and Friedrich-Alexander-Universit\"at Erlangen-Nürnberg, Germany
\and Klinikum rechts der Isar, Technical University of Munich, Germany
\email{c.ouyang@imperial.ac.uk
}}
\maketitle
\begin{abstract}
Probability calibration for deep models is highly desirable in safety-critical applications such as medical imaging. It makes output probabilities of deep networks interpretable, by aligning prediction probability with the actual accuracy in test data. In image segmentation, well-calibrated probabilities allow radiologists to identify regions where model-predicted segmentations are unreliable. These unreliable predictions often occur to out-of-domain (OOD) images that are caused by imaging artifacts or  unseen imaging protocols. Unfortunately, most previous calibration methods for image segmentation perform sub-optimally on OOD images.
To reduce the calibration error when confronted with OOD images, we propose a novel post-hoc calibration model. Our model leverages the pixel susceptibility against perturbations at the local level, and the shape prior information at the global level. The model is tested on cardiac MRI segmentation datasets that contain unseen imaging artifacts and images from an unseen imaging protocol. We demonstrate reduced calibration errors compared with the state-of-the-art calibration algorithm.  
\end{abstract}
\blfootnote{Link to code: \url{https://github.com/cheng-01037/Probability_Calibration_for_OOD_MRI_Segmentation}}

\section{Introduction}

In safety-critical applications like medical imaging, segmentation models are required to produce accurate predictions on clean input data and are also expected to be \textit{aware} of predictions for which the model has \textit{low confidence}, when confronted with out-of-domain (OOD) data. In medical imaging, OOD data is often caused by imaging artifacts or changes in imaging protocols. The awareness of uncertainty allows to alert radiologists about potentially unreliable predictions. Unfortunately, deep models are found to be generally over-confident about predicted probabilities \cite{nguyen2015deep,gonzalez2021detecting}.

\textit{Probability calibration} corrects over- or under-confident predictions, and makes prediction probability \textit{interpretable}, by aligning it with the accuracy on the test dataset. For example, if a segmentation model yields a \textit{confidence} (the probability of the highest-scored class) of 70\% for each pixel in a test image, we say the model to be well-calibrated if 70\% of the pixels are correctly predicted \cite{ding2021local}.  

Unfortunately, most existing probability calibration methods cannot be directly applied to medical image segmentation due to the following reasons: First, the majority of existing methods are designed for image classification, which yield a single class probability per image \cite{platt1999probabilistic,zadrozny2001obtaining,guo2017calibration,tomani2021towards,ji2019bin}. Secondly, most previous methods assume training and testing images are from a same domain. However, we argue that it is the OOD image for which probability calibration is most desirable, while most calibration methods are shown to perform sub-optimally on OOD images \cite{ovadia2019can}. Therefore, in this study we particularly focus on improving calibration for corrupted medical images. 

In this work, we propose a new learning-based probability calibration model for medical image segmentation on out-of-domain (OOD) data. 
Particularly, we focus on the most flexible calibration setting: \textit{post-hoc} calibration that can be applied to various frozen feed-forward networks. Specifically, our calibration model outputs a \textit{tempeature map} that re-adjusts the prediction probability of the segmentation network \cite{guo2017calibration,ding2021local}, correcting over- or under-confident probabilities. Unlike the state-of-the-art method \cite{ding2021local} that only considers the pixel values of input images and their logits, our model finds unreliable predictions by considering how susceptible the prediction of each pixel is, against small perturbations. Such susceptibility helps to reveal the uncertainty caused by the real-world perturbations that originate from imperfect acquisition process (device noise, patient movement \textit{etc.}) or changes in imaging condition (machine vendors, imaging protocols \textit{etc.}). The proposed model further takes advantage of global prior information about the shapes of segmentation targets. These local-level and global-level sources of information strengthens the calibration performance for OOD images. Our contributions can be summarized as follows:
\begin{itemize}
    \item We systematically investigate post-hoc probability calibration for the safety-critical medical image segmentation on out-of-domain (OOD) images.
    \item We propose a new learning-based probability calibration model that incorporates the susceptibility information of pixel-level predictions against perturbations at the local level, and the shape prior information at the global level. The proposed method demonstrates improved performance on OOD testing images compared to the state-of-the-art method. 
    \item We build a comprehensive testing environment for post-hoc calibration, on segmentation for out-of-domain MRI. It incorporates common imaging artifacts: motion artifacts, bias fields, ghosting artifacts, spikes in $k$-space, and an unseen imaging protocol: late gadolinium enhancement (LGE) sequence for MRI. 
\end{itemize}

\section{Related Work}
\noindent \textbf{Probability calibration for image segmentation:} Most probability calibration methods can be categorized into three types: 1) training strategies that intrinsically improve calibration for the task network (classification, regression, \textit{etc.}). These techniques include focal loss \cite{mukhoti2020calibrating}, multi-task learning \cite{karimi2022improving}, adversarial training \cite{kireev2021effectiveness}; 2) Bayesian methods that carefully model the uncertainties of model parameters, input data and/or labeling process \cite{gal2016dropout,kendall2017uncertainties,wang2019aleatoric,mehrtash2020confidence,baumgartner2019phiseg}; 3) post-hoc methods that post-process the softmax output (probability) of an already-trained task network \cite{platt1999probabilistic,zadrozny2001obtaining,guo2017calibration,ding2021local}. Our work follows the post-hoc framework due to its superior flexibility: being applicable to most of already-trained task networks.

More recently, several papers have discussed calibration for image segmentation: \cite{mehrtash2020confidence} evaluates the effects of segmentation losses, model ensembling and MC-dropout on calibration. \cite{karimi2022improving} demonstrates that multi-task learning improves calibration. However, neither works contribute further to post-hoc calibration. Our idea of using data augmentation to estimate susceptibility of pixel-level predictions, which can be interpreted as aleatoric uncertainty estimation, is inspired by \cite{wang2019aleatoric}. However, \cite{wang2019aleatoric} does not investigate post-hoc calibration itself. Our method is built on the state-of-the-art local temperature scaling (LTS) \cite{ding2021local}. To reduce the calibration error on OOD images, we extend LTS by incorporating pixel-level susceptibility and global-level shape prior information.

\noindent \textbf{Segmenting out-of-domain medical images:}
A robust image segmentation model can usually be obtained by applying input-level or feature-level data augmentations \cite{zhang2020generalizing,chen2020realistic,ouyang2021causality}, or by enforcing shape priors \cite{larrazabal2020post,liu2022single,chen2021cooperative}. Unlike these works, our method focuses on the under-explored problem of promoting interpretability of prediction probabilities, especially for those on out-of-domain images. 

\noindent \textbf{Segmentation quality assessment:} Segmentation quality assessment \cite{robinson2017automatic,li2022towards,wang2020deep} predicts a global model performance score, and/or makes corrections to the predicted segmentation labels. Probability calibration is more challenging, as it is required to make continuous, pixel-wise adjustments to prediction probabilities.

\section{Method}
\label{sec: method}
\label{subsec: problem}
\noindent \textbf{Model-based post-hoc calibration:} We aim to align the prediction probability with the accuracy on the test dataset. To this end, in model-based post-hoc calibration, we build a separate calibration model $g_{\phi}(\cdot)$ for a pre-trained task model (in our case segmentation) $f_{\theta}(\cdot)$. To train the calibration model, the validation dataset for the task model is re-used for building $g_{\phi}(\cdot)$. We let $\mbx_i \in \mathbb{R}^{1 \times M \times N}$ be the image, $\mby_i \in \mathbb{R}^{C \times M \times N}$ the ground truth segmentation in the form of one-hot encoding, where $(M,N)$ is the spatial size and $C$ the number of classes. Note, it is usually desirable that the calibration process does not affect the categorical prediction $\hat{\mby}_i$ for segmentation (therefore does not change the accuracy of $f_{\theta}(\cdot)$). 

\noindent \textbf{Temperature scaling:} Temperature scaling \cite{nixon2019measuring,ding2021local} is one of the most simple and effective frameworks for probability calibration. It produces a temperature factor (or map) $\mbT_i>0$ to weigh over-confident predictions down while boost under-confident ones. Formally, let $\mbz_i = f_{\theta}(\mbx_i), \ \mbz_i \in \mathbb{R}^{C \times M \times N}$ be the output logits, let $\sigma(\cdot)$ denote the softmax function along the channel dimension, we naturally have the uncalibrated probability $\hat{\mby}^u_i = \sigma(\mbz_i)$. While with the temperature map $\mbT_i \in \mathbb{R}^{C \times M \times N}$, the calibrated probability $\hat{\mby}^c_i$ can be obtained by re-scaling the logits using $\mbT_i$, \textit{i.e.} $\hat{\mby}^c_i = \sigma(\mbz_i / \mbT_i)$\footnote{To ensure that the calibration does not affect the accuracy of the task network, for each spatial location $(m,n)$ in $\mbT_i$, it is usually assumed that $\mbT_i(c_j, m, n) = \mbT_i(c_k, m, n), \ \forall (c_j, c_k) \in \{1,2,3,...,C\}$, \textit{i.e.}, temperature values remain the same for different channels/classes) \cite{guo2017calibration,ding2021local}.}.

\begin{figure}[!t]
\centering
\includegraphics[width=1.0\linewidth]{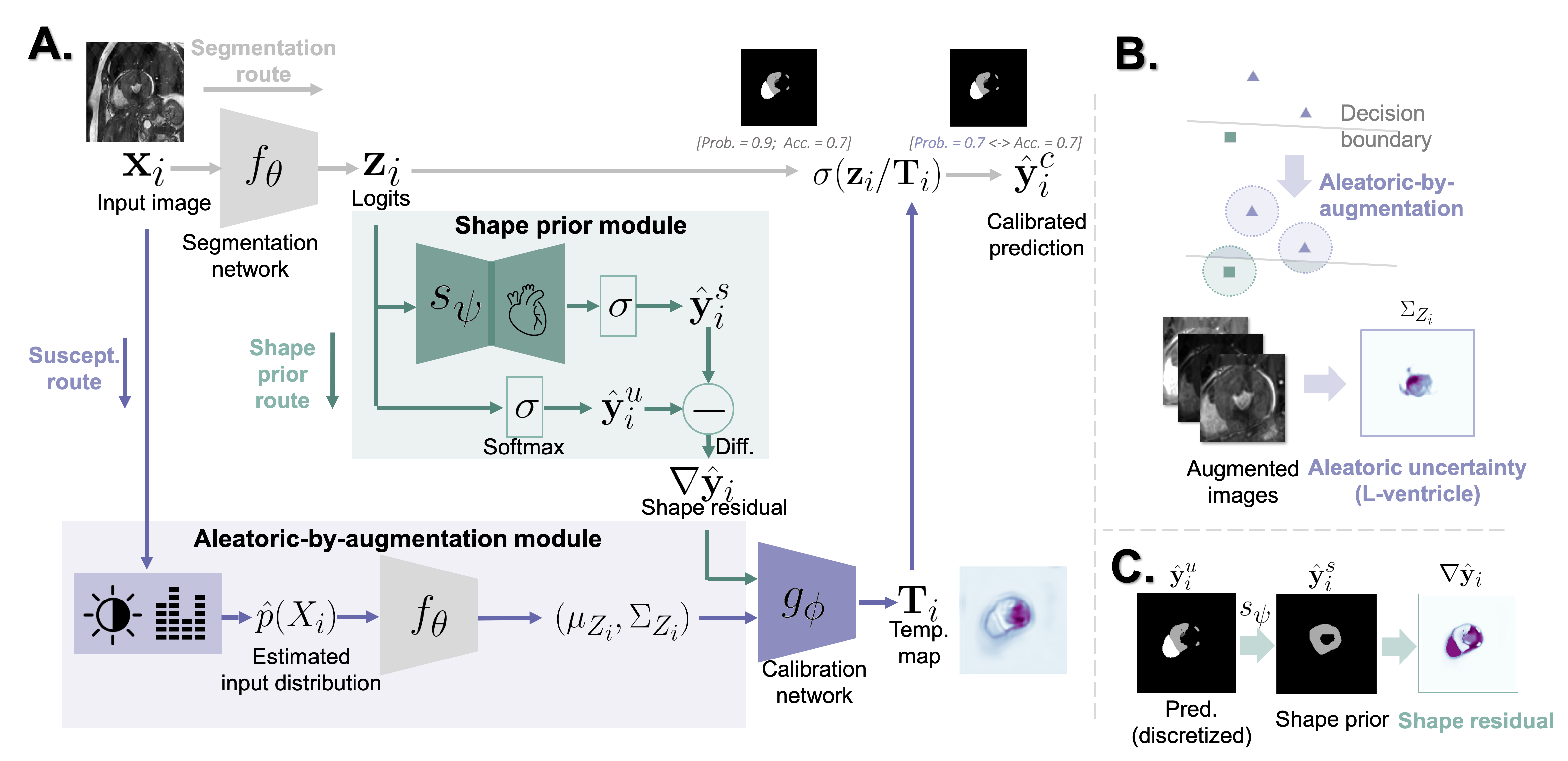}
\caption{\textbf{A.} \textbf{Workflow of the proposed calibration technique:} A temperate map $\mbT_i$ is used to adjust probabilities of a segmentation network. To do this, the image $\mbx_i$ is sent through a \textcolor{gray}{segmentation network} $f_{\theta}(\cdot)$ to obtain the logits $\mbz_i$. Meanwhile, to obtain $\mbT_i$, $\mbx_i$ is sent through two calibration routes: In the \textcolor{violet}{susceptibility route}, the estimated distribution $\hat{p}(X_i)$ of $\mbx_i$ is obtained by repeated data augmentations. The uncertainty $(\mu_{Z_i}, \Sigma_{Z_i})$ is computed by sending samples of $\hat{p}(X_i)$ to $f_{\theta}(\cdot)$. In the \textcolor{teal}{shape prior route}, $\mbz_i$ is sent to the shape prior network $s_{\psi}(\cdot)$ to obtain a shape residual $\nabla \hat{\mby}_i$ which highlights the regions where the prediction differs from the prior knowledge about plausible shapes of segmentation targets. The calibration network $g_{\phi}(\cdot)$ takes $(\mu_{Z_i}, \Sigma_{Z_i})$ and $\nabla \hat{\mby}_i$ as inputs and estimates $\mbT_i$ for rescaling logits $\mbz_i$ of the segmentation. \textbf{B. Aleatoric uncertainty}  reflects the susceptibility (shaded regions trespassing the decision boundary) of a prediction under small perturbations. \textbf{C. Shape prior and shape residual}, highlighting potentially unreliable predictions.}
\label{fig: overview}
\end{figure}
\noindent \textbf{Method overview:}
We aim to obtain a temperature-scaling-based calibration network $g_{\phi}(\cdot)$ that is suitable for out-of-domain (OOD) testing images. Examples of these OOD images are assumed to be \textit{unseen} by the segmentation network $f_{\theta}(\cdot)$ and the calibration network $g_{\phi}(\cdot)$ during their training processes. To this end we propose to 1) provide the susceptibility of the prediction of each pixel against small perturbations caused by potential image corruption/artifact or a change in imaging protocol. This susceptibility reflects how likely the prediction of a pixel might be altered when real image artifacts or changes in imaging protocols are present. This is also known as \textit{aleatoric uncertainty} \footnote{We do not explicitly highlight it as aleatoric uncertainty, since we do not have the ground truth to evaluate the accuracy of this estimation of aleatoric uncertainty.} \cite{kendall2017uncertainties,wang2019aleatoric}. 2) We also provide the calibration network with prior information about the shape of the target segmentation. This shape prior is encoded by a denoising autoencoder $s_{\psi}(\cdot)$ and it provides a second opinion about the correctness of the prediction. 

As shown in Fig.~\ref{fig: overview}\textcolor{red}{-A}, to obtain the temperature map $\mbT_i$, the input $\mbx_i$ is fed to two modules: The \textit{Aleatoric-by-augmentation} module (colored in \textcolor{violet}{purple}) estimates the pixel-level susceptibility (aleatoric uncertainty) by repeated data augmentations. The \textit{shape prior module} (colored in \textcolor{teal}{green}) compares the uncalibrated prediction with the shape prior encoded in the denoising autoencoder $s_{\psi}(\cdot)$, and provides the calibration network $g_{\phi}(\cdot)$ with the residual between the uncalibrated prediction and the prior. The calibration network $g_{\phi}(\cdot)$ takes the outputs of the two modules, and estimates a temperature map for adjusting $\mbz_i$. Finally, the calibrated prediction is made by passing $\mbz_i / \mbT_i$ to a softmax layer. 

\noindent \textbf{Aleatoric uncertainty by augmentation:} The aleatoric-by-augmentation module provides the calibration network $g_{\phi}(\cdot)$ with information about susceptibility of predictions for each pixel against small perturbations. Intuitively, if the prediction can be easily flipped by a small perturbation, the prediction of that pixel could be unreliable. In medical images, OOD images can also be viewed as being generated by perturbing intra-domain images \cite{chen2021cooperative}.

To formally model this susceptibility, we resort to the concept of \textit{aleatoric uncertainty} \cite{gal2016dropout,kendall2017uncertainties,wang2019aleatoric}. As shown in Fig. \ref{fig: overview}\textcolor{red}{-B}, it models images to have a distribution $p(X_i)$ arising from the acquisition process, rather than treating each image as a single data point (which is instead assumed by the state-of-the-art LTS \cite{ding2021local}).
This modeled distribution can be written as
$p(X_i) = \int p(X_i | a)p(a)da$,
where $p(X_i | a)$ represents the image acquisition process and $a \sim p(A)$ denotes the ``randomness'' within different possible acquisition processes \cite{wang2019aleatoric}.
Then, the susceptibility (uncertainty) can be estimated by propagating $p(X_i)$ through the segmentation model $f_{\theta}(\cdot)$.

In practice, inspired by \cite{wang2019aleatoric,ouyang2021causality}, we employ data augmentation to obtain the estimation $\hat{p}(X_i)$ of the real $p(X_i)$. Specifically, for each image $\mbx_i$, we perform repeated augmentations to obtain $\{ \mbx_{i,l}' | \mbx_{i,l}' = \mathcal{T}_{a_{l}'}(\mbx_i), \ a_{l}' \sim p(A') \}$, where $l = 1,2,3, ..., N_A$ is the index of augmented samples and $\mathcal{T}_{a_{l}'}(\cdot)$'s are photometric augmentations parameterized by $a_{l}'$'s. To ensure fairness,
$\mathcal{T}_{a_{l}'}(\cdot)$'s are configured to be the \textit{same} types of photometric augmentations used for training $f_{\theta}(\cdot)$ and they \textit{do not} incorporate the corruptions (artifacts) in the testing data. Then, the propagated uncertainty in the logits, in the form of mean $\mu_{Z_i}$ and variance $\Sigma_{Z_i}$, can be computed by sending $\{ \mbx_{i,l}' \}$ to the segmentation network $f_{\theta}(\cdot)$. For simplicity, when computing $\Sigma_{Z_i}$, each pixel is assumed to be independent.

\noindent \textbf{Shape prior:}
To provide a second opinion about the correctness of the segmentation, shape priors~\cite{robinson2017automatic,raju2021deep,larrazabal2020post} are used. For probability calibration, if the predicted shape deviates largely from the prior information about plausible shapes, the prediction is likely to be unreliable.

Here, we employ a denoising autoencoder as the shape prior model. It memorizes correct shapes of segmentation targets in the validation dataset. As shown in the \textcolor{teal}{green} block in Fig. \ref{fig: overview}\textcolor{red}{-A}, the autoencoder $s_{\psi}(\cdot)$ takes the uncalibrated logits $\mbz_i$ as the input and produces a denoised plausible shape $\hat{\mby}_i^s$ of the segmentation target, in the form of probabilities. To highlight regions where the uncalibrated prediction $\hat{\mby}^u_i = \sigma(\mbz_i)$ deviates from the plausible shape $\hat{\mby}_i^s$, we send the shape residual $\nabla \hat{\mby}_i =  \hat{\mby}_i^s - \hat{\mby}^u_i  $ to the calibration network. An example of a shape residual is shown in Fig. \ref{fig: overview}\textcolor{red}{-C}. 
In practice, to avoid learning an identity mapping, we apply heavy dropout to the encoder part of $s_{\psi}(\cdot)$ during training.

Unlike the shape priors in \cite{larrazabal2020post,chen2021cooperative}, which directly correct the prediction, we do not expect $s_{\psi}(\cdot)$ to provide highly accurate segmentations: As shown in Fig. \ref{fig: overview}\textcolor{red}{-C}., the right ventricle has been correctly predicted by $f_{\theta}(\cdot)$ while $s_{\psi}(\cdot)$ (erroneously) disagrees. Instead, we only expect the shape prior module to highlight potentially implausible regions. We leave the calibration network to make the final decision.

\noindent \textbf{Calibration network:}
The calibration network $g_{\phi}(\cdot)$ produces a temperature map $\mbT_i$ that is specific to $\mbx_i$, by considering the pixel-level susceptibility (uncertainty) $(\mu_{Z_i}, \Sigma_{Z_i})$  and the shape residual $\nabla \hat{\mby}_i$. Following the baseline LTS \cite{ding2021local}, we also send the image $\mbx_i$ and the uncalibrated logits $\mbz_i$ to $g_{\phi}(\cdot)$. After $\mbT_i$ is computed, the calibrated prediction $\hat{\mby}^c_i$ is given by $
    \hat{\mby}^c_i = \sigma(\mbz_i /  \mbT_i), \text{ where } \mbT_i = g_{\phi}(\mu_{Z_i}, \Sigma_{Z_i}, \nabla \hat{\mby}_i, \mbz_i, \mbx_i)$, and $\sigma(\cdot)$ denotes the softmax layer.

In practice, we configure $g_{\phi}(\cdot)$ as a shallow residual network, which we empirically found to yield comparable results to the decision-tree-inspired network in the vanilla LTS~\cite{ding2021local}, but to be more flexible in terms of model architecture. A channel attention layer is used in $g_{\phi}(\cdot)$ to allow the network to adaptively weigh information from different sources.

\noindent \textbf{Training objectives}:
Following the standard setting of post-hoc calibration, both the calibration network $g_{\phi}(\cdot)$ and the shape prior module $s_{\psi}(\cdot)$ are trained on the validation dataset used in building the segmentation network $f_{\theta}(\cdot)$. To avoid shortcut learning from $s_{\psi}(\cdot)$ to $g_{\phi}(\cdot)$, two networks are trained one by one. We first train the shape prior module using the cross entropy loss:
\begin{align}
     \small
     \centering
    \mathcal{L}_{s}(\psi) = - \frac{1}{MN} \sum_{m,n} \sum_c \mby_i(c,m,n) \log \big(\sigma(s_{\psi}(\mbz_i(c,m,n))) \big), 
    \label{equ: shape}
\end{align}
where $\mbz_i = f_{\theta}(\mbx_i), \ (\mbx_i, \mby_i) \in \mathcal{D}_{val}$ the validation set, the subscript $c$ denotes the class index.
After $s_{\psi}(\cdot)$ is trained, we close the gradient computation for $s_{\psi}(\cdot)$. We then train the calibration network $g_{\phi}(\cdot)$ using the negative log likelihood loss that is commonly used for training post-hoc calibration networks \cite{guo2017calibration,ding2021local,nixon2019measuring}:
\begin{align}
    \small
    \centering
    \mathcal{L}_g (\phi) = - \frac{1}{MN} \sum_{m,n} \sum_c \mby_i(c,m,n) \log \big( \sigma(\mbz_i(c,m,n) / \mbT_i(c, m,n))  \big),
    \label{eq: calib_loss}
\end{align}
where $\mbT_i = g_{\phi}(\mu_{Z_i}, \Sigma_{Z_i},\nabla \hat{\mby}_i,  \mbz_i, \mbx_i)$, $(\mu_{Z_i}, \Sigma_{Z_i} )$'s are obtained by sending multiple augmented versions of $\mbx_i$ to $f_{\theta}(\cdot)$. This loss penalizes over-confident erroneous predictions while it encourages high confidence for correct predictions. Although Eq.~\ref{eq: calib_loss} has similar form as cross-entropy, it essentially optimizes over $\phi$ via $\mbT_i$. Since at each location $(m,n)$, $\mbT_i(c, m,n)$'s remain constant for all the classes $c$'s, this loss does not affect the categorical segmentation result~\cite{guo2017calibration,ding2021local}.

\section{Evaluation and Results}

\begin{table}[!t]
\caption{Quantitative results on expected calibration error (ECE) and static calibration error (SCE). Lower the better. Average Dice scores of the segmentation networks are appended for reference.}
    \centering
    \resizebox{1.0\linewidth}{!}{%
    \begin{tabular}{c|c|cccc|c|c|c|cccc|c|c}
    \toprule
    \multirow{2}{*}{Method}   & \multicolumn{7}{c|}{ECE [\%] $\downarrow$}  & \multicolumn{7}{c}{SCE [\%] $\downarrow$}  \\
    & \multicolumn{1}{c}{Intra-dom.} & Bias field & Motion & Ghosting & \multicolumn{1}{c}{Spike} & \multicolumn{1}{c}{Artifact Avg.} & Cross Seq. & \multicolumn{1}{c}{Intra-dom.} & Bias field & Motion & Ghosting & \multicolumn{1}{c}{Spike} & \multicolumn{1}{c}{Artifact Avg.}  & Cross Seq.\\
    \hline
    
    UC  & 10.29  & 13.60 & 22.38 & 19.68 & 39.05 & 23.67 & 30.29 & 5.27 & 6.96 & 11.45 & 10.09 & 19.82 & 12.08 & 15.58 \\
    Alea. \cite{wang2019aleatoric,kendall2017uncertainties} &  7.74 & 9.06 & 16.47 & 16.78 & 37.31 & 19.90 & 28.50 &5.08 & 8.39 & 10.11 & 10.56 & 21.15 & 12.55 & 16.89 \\
    TS \cite{guo2017calibration} & 10.06 & 13.31 & 22.04 &  19.42 & 38.87 & 23.41 & 29.96 & 5.17 & 6.84 & 11.31 & 9.99 & 19.76 & 11.98 & 15.47 \\
    LTS \cite{ding2021local} & 3.22 & 5.46  & 10.21 & 10.61 & 31.60 & 14.48 & 16.78 & 3.63 & 4.91 & 7.93 & 7.80 & 17.80 & 9.61& 11.52 \\
    \hline
     \multirow{2}{*}{Proposed} & \textbf{3.12} & \textbf{4.65}$^{*}$ & \textbf{8.88}$^{\ddagger}$ & \textbf{9.23}$^{*}$ & \textbf{28.35}$^{\dagger}$ & \textbf{12.78} & \textbf{15.37}$^{\dagger}$ &\textbf{3.38} & \textbf{4.75}$^{*}$ & \textbf{7.23}$^{\dagger}$ & \textbf{7.14}$^{\ddagger}$ & \textbf{16.45}$^{\ddagger}$ & \textbf{8.89} & \textbf{10.77}$^{\dagger}$ \\
     & (-0.10) & (-0.82) & (-1.33) & (-1.38) & (-3.26) & (-1.70) & (-1.41) & (-0.25)
    & (-0.16) & (-0.70) & (-0.67) & (-1.35) & (-0.72) & (-0.75)\\
    \bottomrule 
    \multicolumn{8}{l}{
    \scriptsize
      $^{\dagger}$: $p$-value $<$ 0.01; $^{\ddagger}$: $p$-value $<$ 0.05; $^{*}$: $p$-value $>$ 0.05, compared with the results of LTS \cite{ding2021local}.
      }\\
    \cline{1-8}
    \multirow{2}{*}{} & \multicolumn{7}{c}{Dice score [\%] $\uparrow$} \\
    & Intra-dom. & Bias field & Motion & Ghosting & \multicolumn{1}{c}{Spike} & \multicolumn{1}{c}{Artifact Avg.} & \multicolumn{1}{c}{Cross Seq.}  \\
    \cline{1-8}
     Seg. Net. & 85.14 & 80.29 & 69.02 & 79.73 & 39.02 & 67.02 & \multicolumn{1}{c}{62.74}\\
    \cline{1-8}
    \end{tabular}}
    \label{tbl: performance}
\end{table}

\begin{table}[h]

\caption{Ablating key components and the number of test-time augmentations, evaluated on artifact-corrupted images.}
\centering
\resizebox{0.5\linewidth}{!}{%
\begin{tabular}{cc|cc|c|cc} 
\toprule
Alea. & Shape & ECE [\%] $\downarrow$ & SCE [\%] $\downarrow$ & No. of Aug. & ECE [\%] $\downarrow$ & SCE [\%] $\downarrow$\\ 
\hline
$\times$ & $\times$ & 14.48 & 9.61 &  15 & 13.22 & 9.00  \\
\checkmark & $\times$ & 13.03 & 9.20  &  45 & 12.94 & 8.94  \\
$\times$ & \checkmark & 13.50 & 9.18 &  90 & 12.85 & 8.92 \\
\hline
\checkmark & $\checkmark$ & 12.78 & 8.89 &  180 & 12.78 & 8.89  \\
\bottomrule
\end{tabular}
}
\label{tbl: abl_joint}
\end{table}
\noindent \textbf{Dataset:}
\noindent \textit{Training and validation dataset}: We employ the ACDC cardiac MRI segmentation dataset (bSSFP sequence) \cite{bernard2018deep} for building the segmentation model and the proposed calibration model. Specifically, we take the ES fold of ACDC and split it into training, validation and (intra-domain) testing sets of 60/20/20 cases. To simulate data-hungry medical image segmentation \cite{chen2021cooperative}, each time we take 20 cases out of the training data for building the segmentation network, and 5 out of validation data for validating the segmentation network and for training the calibration model. We repeat this process for 3 times to cover all the training samples, and obtain 3 segmentation models. For each segmentation model, we repeat training the calibration model for 3 times.

\noindent \textit{Artifact-corrupted testing dataset}: Inspired by \cite{chen2021cooperative}, we simulate common MRI artifacts: bias field, motion artifact, ghosting artifact and $k$-space spikes, separately, to the 20 intra-domain testing cases mentioned above, using TorchIO \cite{perez2021torchio}. Using this controlled environment allows us to observe the model behaviors under each type of artifacts.

\noindent \textit{Cross-sequence testing dataset}: We further test the above ACDC-based models on the 40 LGE MRI of the testing fold of the MS-CMRSeg challenge \cite{zhuang2022cardiac}. As ACDC is based on bSSFP sequence, the segmentation and calibration models have never seen images from LGE sequence before testing.

\noindent \textbf{Network architecture and training configurations:} We employ a U-Net~\cite{ronneberger2015u} as the segmentation network. For the calibration network $g_{\phi}(\cdot)$, we employ a shallow ResNet with 5 input branches for processing $\nabla \hat{\mby}_i$, $\mu_{Z_i}$, $\Sigma_{Z_i}$, $\mbz_i$, and $\mbx_i$ separately. 
These branches are merged by a channel attention block, followed by two ResNet blocks. The shape prior model $s_{\psi}(\cdot)$ is configured as a small U-Net with dropout ($p$=0.5) in its encoder. The Adam optimizer is used, with an initial learning rate of $1\times 10^{-3}$, 800 epoches separately for $s_{\psi}(\cdot)$ and $g_{\phi}(\cdot)$. 
In each iteration, $(\mu_{Z_i}, \Sigma_{Z_i})$ are computed by repeating augmentations for 6 times.

Photometric transforms: brightness, contrast, gamma transform, random additive noises \cite{chen2021cooperative}, and geometric transformations: affine transformation and elastic transformation are used as data augmentation for training the segmentation model and the calibration model (also for the  LTS~\cite{ding2021local}). Importantly, these data augmentations \textit{do not} include the corruptions in the testing data.

\begin{figure}[!t]
\centering
\includegraphics[width=1.0\linewidth]{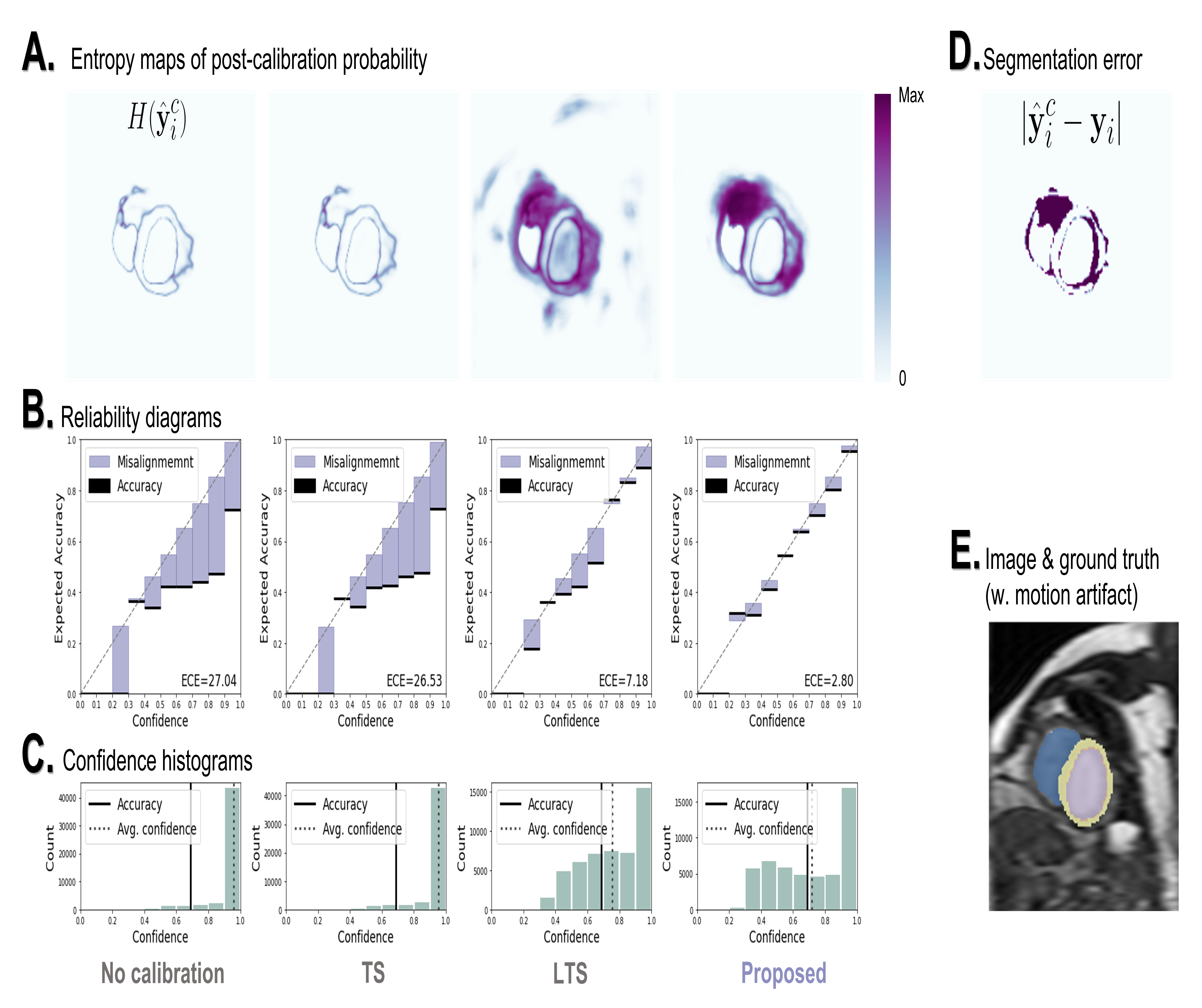}
\caption{\textbf{A.} For the proposed method, the entropy map which shows the doubt of the calibration model, agrees well with the actual segmentation error (shown in \textbf{D.}). \textbf{B.} Reliability map of the proposed method demostrates the least misalignment (\textcolor{violet}{purple bars}) between confidence and accuracy. \textbf{C.} The confidence histogram shows that the proposed method has corrected over-confident predictions, compared with uncalibrated results. \textbf{E.} The motion-corrupted input image and its ground truth segmentation.}
\label{fig: results}
\end{figure}

\noindent \textbf{Quantitative and qualitative results:} We employ commonly-used expected calibration error (ECE) \cite{naeini2015obtaining} and static calibration error (SCE) \cite{nixon2019measuring} for evaluation (lower the better). Both of them measure the gap between prediction probability and the accuracy in test time, and the latter is a class-conditional version of the former. To account for the foreground-background class imbalance in ACDC, inspired by  \cite{mehrtash2020confidence}, these two metrics are computed over the region-of-interests obtained by dilating (expanding) the ground truth segmentations with a kernel size of 10 pxiels.

As shown in Table \ref{tbl: performance}, we compare the proposed method with the uncalibrated model (UC) and the state-of-the-art local temperature scaling (LTS) \cite{ding2021local}. The proposed method demonstrates overall smaller calibration errors compared with LTS. Calibration errors of the estimated aleatoric uncertainty (Alea.) \cite{kendall2017uncertainties,wang2019aleatoric} and temperature scaling (TS) \cite{guo2017calibration} are also presented. The segmentation performances measured in Dice scores of the segmentation networks are also attached.

We show in the first row of Fig. \ref{fig: results} the entropy maps $H(\hat{\mby}^c_i)$'s of the calibrated probabilities, where higher values suggest stronger doubts by the calibration network. The entropy map produced by the proposed method has the best agreement with the actual segmentation error. We further show the reliability map in the second row, where the \textcolor{violet}{purple bars} represent the gaps between confidence (x-axis) and accuracy (y-axis) at each confidence level. The proposed method also yields the smallest gaps. Confidence histograms of post-calibration probabilities are shown in the third row.

\noindent \textbf{Ablation studies:} We ablate the two key components of the proposed method: the susceptibility (aleatoric uncertainty) estimation and the shape prior. The results in Table \ref{tbl: abl_joint} left show that the best performances are obtained when two components work together. We also ablate the number of repeated augmentations $N_A$ used for estimating susceptibility during \textit{test time}. As shown in Table~\ref{tbl: abl_joint} right, a larger $N_A$ leads to more precise estimations, yielding less errors.

\noindent\textbf{Conclusion:}
In this work we propose a new calibration method for out-of-domain MRI segmentation. 
Future works can be done by designing better shape prior models that can account for segmentation targets with more irregular shapes, like blood vessels and tumors.

\noindent\textbf{Acknowledgments:} This work was in part supported by EPSRC Programme Grants (EP/P001009/1,
EP/W01842X/1) and in part by the UKRI London Medical Imaging and Artificial Intelligence Centre for Value Based Healthcare (No.104691). S.W. was also supported by the Shanghai Sailing Programs of Shanghai Municipal Science and Technology Committee (22YF1409300).

%
\bibliographystyle{splncs}
\bibliography{main.bbl}
\end{document}